\documentclass{article}

\usepackage{PRIMEarxiv}

\usepackage[english]{babel}
\usepackage{microtype}
\usepackage{booktabs}
\usepackage{etoolbox}

\usepackage[T1]{fontenc}

\usepackage{graphicx}
\usepackage{diagbox}
\usepackage{amsmath}
\usepackage{subcaption}
\usepackage{hyperref}
\usepackage{siunitx}

\title{Molecular Machine Learning Using Euler Characteristic Transforms
}

\author{ \href{https://orcid.org/0009-0006-1316-9026}{\includegraphics[scale=0.06]{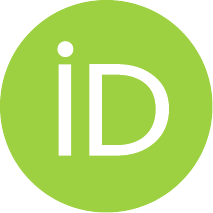}}Victor Toscano-Duran\thanks{Corresponding author.} \\
  Department of Applied Mathematics I, University of Seville \\
  Seville, Spain \\
  \texttt{vtoscano@us.es} \\
   \And
  \href{https://orcid.org/0009-0001-4718-9168}{\includegraphics[scale=0.06]{orcid.pdf}}Florian Rottach  \\
  Central Data Science, Boehringer Ingelheim GmbH \\
  Biberach/Riss, Germany \\
  School of Medicine, University of Tübingen \\
  Tübingen, Germany \\
  \texttt{florian.rottach@boehringer-ingelheim.com} \\
  \And
  \href{https://orcid.org/0000-0003-4335-0302}{\includegraphics[scale=0.06]{orcid.pdf}}Bastian Rieck \\
  AIDOS Lab, University of Fribourg \\
  Fribourg, Switzerland \\
  \texttt{bastian.grossenbacher@unifr.ch}
}

\begin{document}
\maketitle



\begin{abstract} 
The shape of a molecule determines its physicochemical and biological properties. However, it is often underrepresented in standard molecular representation learning approaches. Here, we propose using the \emph{Euler Characteristic Transform}~(ECT) as a geometrical-topological  descriptor. Computed directly on a molecular graph derived from handcrafted atomic features, the ECT enables the extraction of \emph{multiscale} structural features,  offering a novel way to represent and encode molecular shape in the feature space.
We assess the predictive performance of this representation across nine benchmark regression datasets, all centered around predicting the inhibition constant $K_i$.
In addition, we compare our proposed ECT-based representation against traditional molecular representations and methods, such as molecular fingerprints/descriptors  and graph neural networks (GNNs). Our results show that our ECT-based representation achieves \emph{competitive performance}, ranking among the best-performing methods on several datasets. More importantly, its combination with traditional representations, particularly with the AVALON fingerprint, significantly \emph{enhances predictive performance}, outperforming other methods on most datasets. 
These findings highlight the complementary value of \emph{multiscale} topological information and its potential for being combined with established techniques. Our study suggests that hybrid approaches incorporating explicit shape information can lead to more informative and robust molecular representations, enhancing and opening new avenues in molecular machine learning tasks. To support reproducibility and foster open biomedical research, we provide open access to all experiments and code used in this work.


\end{abstract}

\section{Introduction}\label{sec:introduction}

Understanding and predicting the properties of molecules is at the core of modern medicine and healthcare. Almost all pharmaceutical agents, like small-molecule drugs or diagnostic agents, are molecular in nature, and their therapeutic efficacy is critically dependent on their physicochemical and biological properties. From drug discovery to personalized medicine, the ability to predict how a molecule will behave in the body (its solubility, toxicity, bioavailability, and binding affinity), as well as predicting molecule properties, is essential for developing safe and effective treatments. Given the high costs and long timelines associated with traditional drug discovery, predictive modeling has emerged as a vital tool for improving the efficiency of the early stages of drug development, enabling faster screening of potential candidates and reducing the need for extensive experimental testing \cite{carracedo2021review,deng2022artificial,walters2020applications}.
However, despite significant advancements, \emph{predicting} molecular properties remains a challenge due to the complexity of molecular behavior. Traditional molecular representations, such as fingerprints or descriptors, often fail to fully capture the rich structure and shape information that influence a molecule’s function in a biological context \cite{david2020molecular,wigh2022review}. This gap is particularly pronounced when trying to predict properties like the inhibition constant ($K_i$), a key parameter in drug--target interactions, where even small changes in molecular shape can lead to large differences in binding affinity \cite{deng2023systematic,van2022exposing}.
Recent developments in molecular machine learning have demonstrated the potential of deep learning techniques in general and graph-based models in particular to improve predictive performance \cite{atz2021geometric,jiang2021could}. However, many of these models still fail to fully capture the topological and geometric aspects of molecular shape, proving detrimental to the learning outcome.
For example, methods like Graph Neural Networks (GNNs) \cite{graphneuralnetworksbook,ZHOU202057}  may capture \emph{some} of the structural information, but they often overlook the importance of representing the shape in a comprehensive and multiscale manner~\cite{Koke25b}.

This is where algebraic topology \cite{hatcher2005algebraic,munkres2018elements} and topological data analysis \cite{edelsbrunner2022computational,lum2013extracting} provide powerful tools to analyze the shape of molecular data, as shown by recent studies \cite{rottach2025topology}. The \emph{Euler Characteristic Transform} (ECT)~\cite{Rieck24a,munch2025invitation}, for example, could be a topological representation of molecules that captures the molecular shape by tracking changes in the Euler characteristic \cite{leinster2008euler} across different scales and directions of the feature space. By incorporating topological shape information, the ECT allows us to represent and encode the molecular structure in a multiscale and novel way, capturing essential shape details that traditional representations may miss. Recent studies have shown the promise of the ECT in other fields, such as biology and material science, where shape plays a crucial role in determining functional properties \cite{ECTapplicationThreshold,Roell24a}.
In this work, we compare and explore the use of the ECT-based molecular representation, which is computed directly over molecular graphs derived from handcrafted atomic features, to predict $K_i$, a key molecular property, as well as perform a comparison between our approach and traditional methods over a series of nine binding affinity datasets. By using this topological representation, as well as combining it with traditional molecular representations, we aim to enhance predictive performance and provide new insights into the role of molecular shape in molecular learning. Our experiments shows that our ECT-based approach exhibites competitive predictive performance, in some cases even outperforming all alternative methods. In addition, our experiments show that the \emph{combination} of our ECT-based approach with existing methods, more specifically with the AVALON fingerprint, leads to improved performance, thus
highlighting the complementary value of multiscale topological and shape information. Ultimately, our work contributes to the growing body of evidence suggesting that incorporating molecular shape at a fundamental level can lead to more robust and informative models, opening up new avenues for the design of better molecular machine learning and more effective therapies.


The remainder of this paper is organized as follows: Firstly, Section \ref{sec:background}  introduces the foundational concepts relevant to our study, including molecular structures and their representations, traditional molecular descriptors, and the Euler Characteristic Transform. Section \ref{sec:materials} details the datasets employed, the traditional representations included for comparison in the experiments, our proposed ECT-based approach, and the overall  experimental setup. The results of our experiments, with a comprehensive discussion, are presented in Section \ref{sec:results}. Finally, conclusions and future work are discussed in Section \ref{sec:conclusion}.

\section{Background}\label{sec:background}

Molecules \cite{corey2007molecules} are the fundamental building blocks of chemical and biological systems. In the context of drug discovery \cite{drews2000drug}, small molecules are designed or screened for their ability to modulate the activity of specific biological targets, typically proteins. The interaction between a drug and its target is governed by a complex interplay of properties, including molecular shape, electronic distribution, and physicochemical characteristics such as hydrophobicity, polarity, and charge. Accurately modeling and predicting these properties is a central challenge in cheminformatics and molecular machine learning \cite{wigh2022review}.

\begin{figure}
\centering
\begin{subfigure}{0.45\textwidth}
    \includegraphics[height=0.7\textwidth]{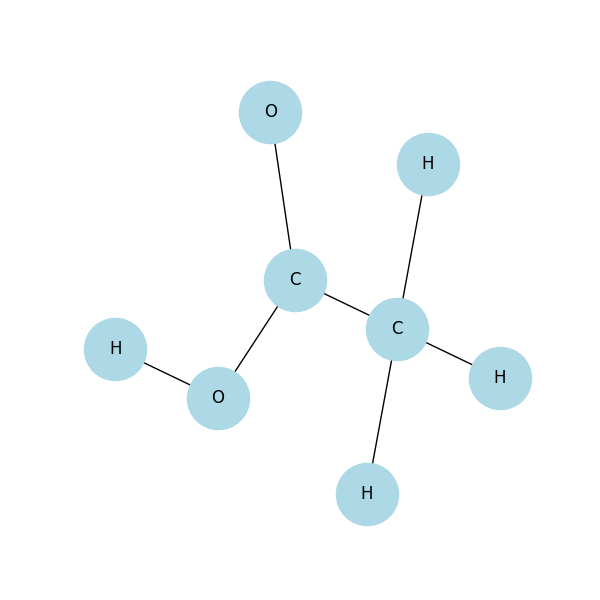}
    \caption{Molecule Graph example.}
    \label{fig:moleculegraph}
\end{subfigure}
\hfill
\begin{subfigure}{0.45\textwidth}
    \includegraphics[height=0.7\textwidth]{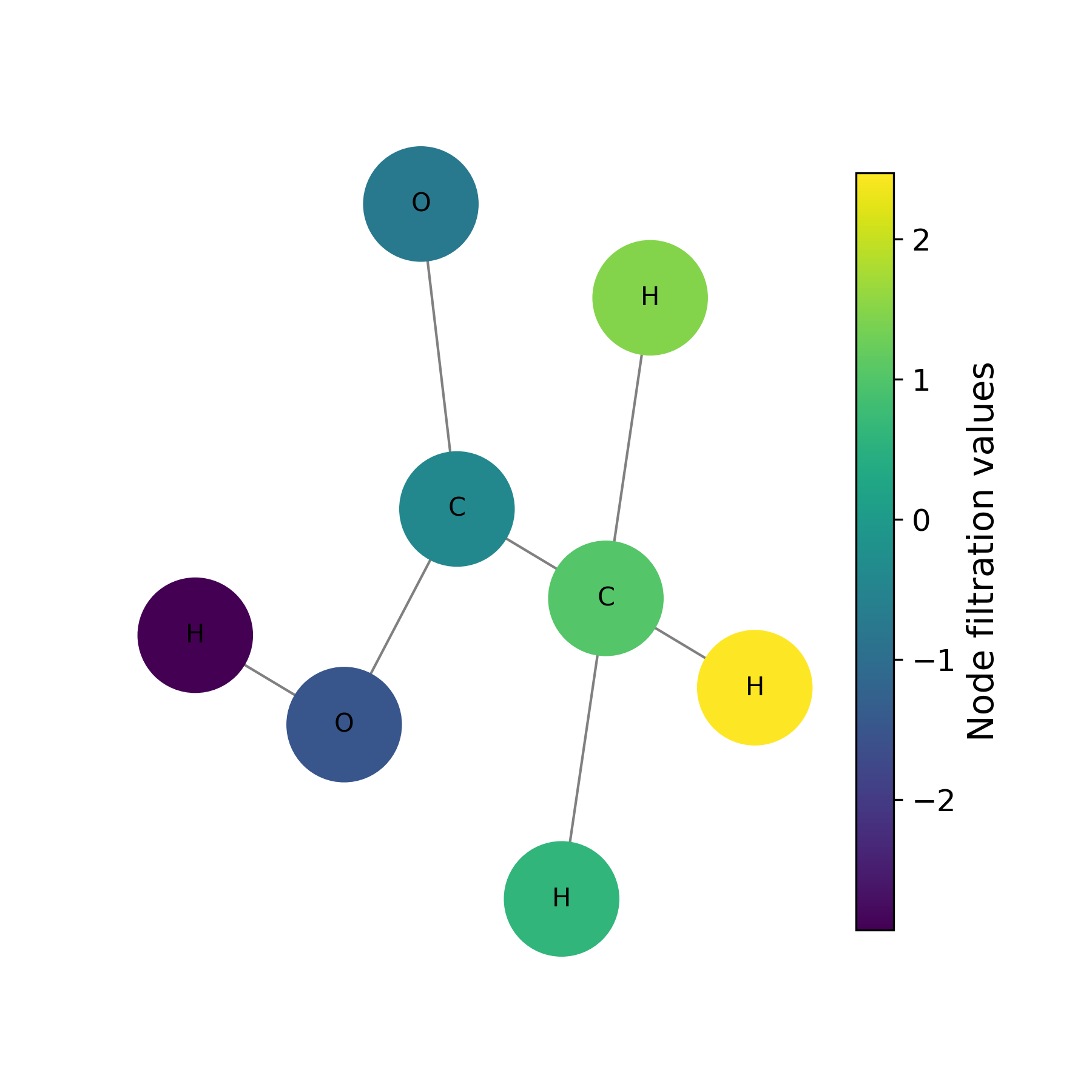}
    \caption{Filtration example}
    \label{fig:filtrationvalues}
\end{subfigure}

\begin{subfigure}{\textwidth}
    \includegraphics[width=0.9\textwidth]{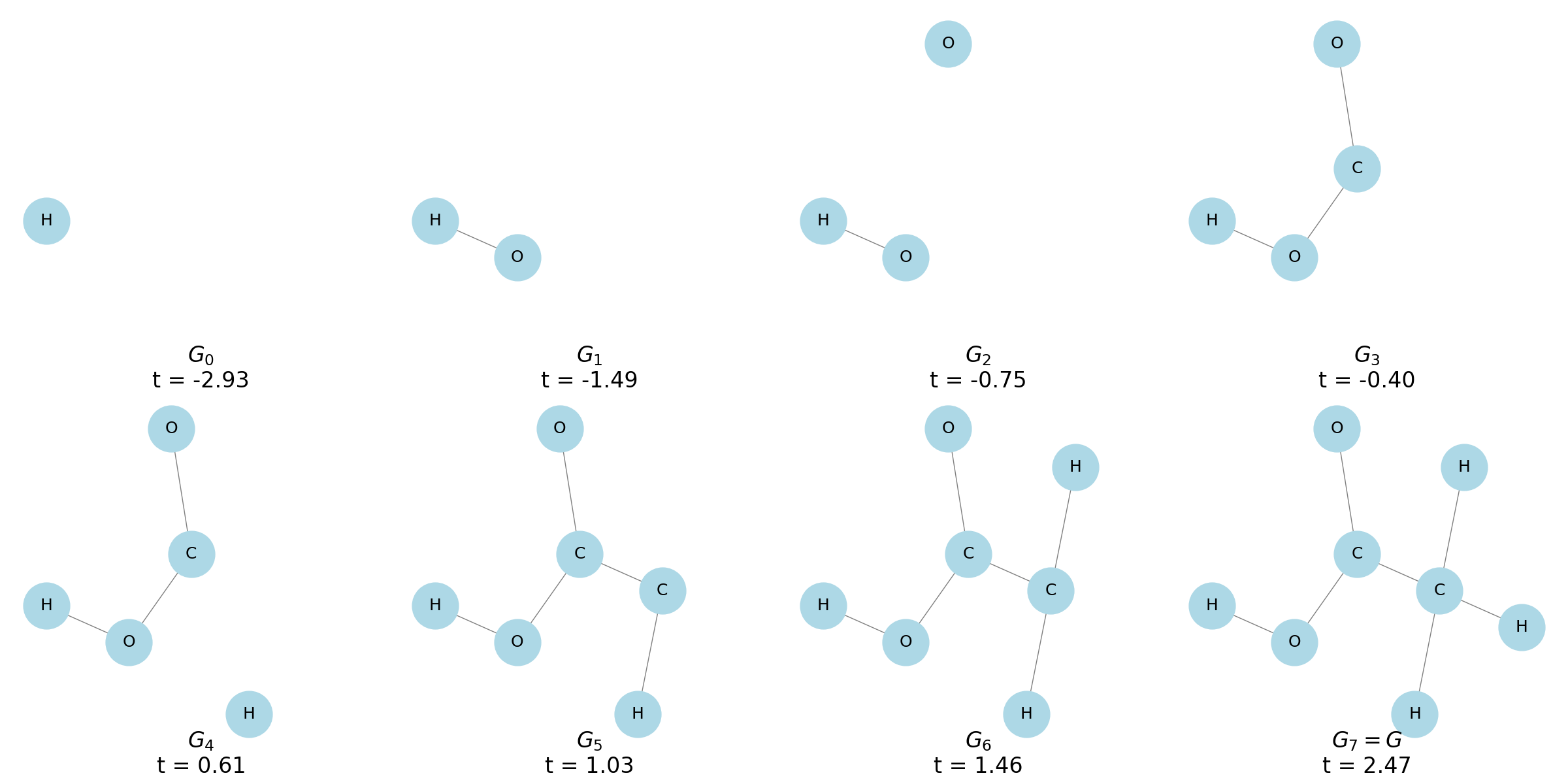}
    \caption{Filtration sequence}
    \label{fig:filtrationsequence}
\end{subfigure}

\begin{subfigure}{0.45\textwidth}
    \includegraphics[width=\textwidth,height=0.7\textwidth]{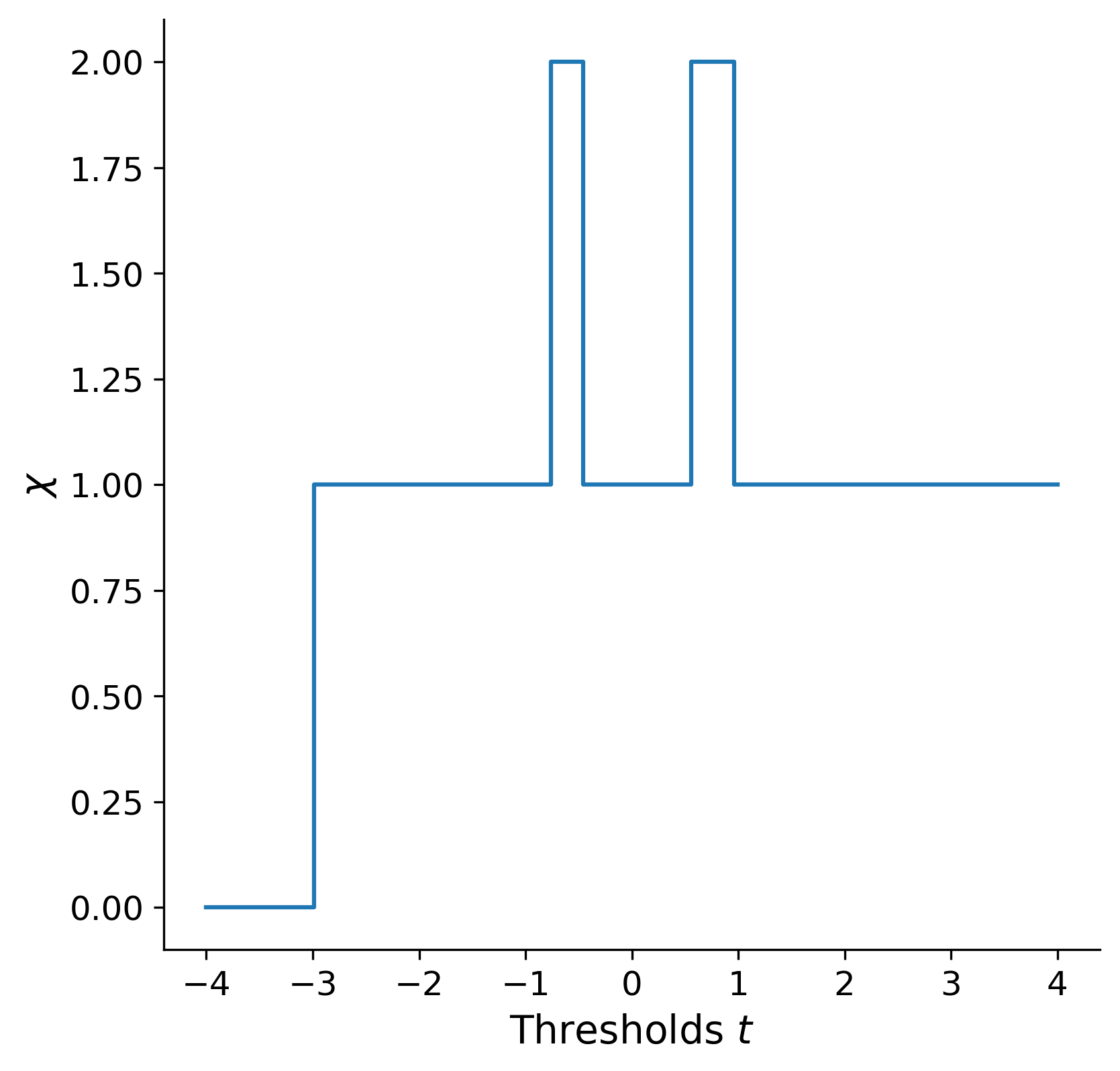}
    \caption{Euler Characteristic Curve}
    \label{fig:eccs}
\end{subfigure}
\hfill
\begin{subfigure}{0.45\textwidth}
    \includegraphics[width=\textwidth,height=0.7\textwidth]{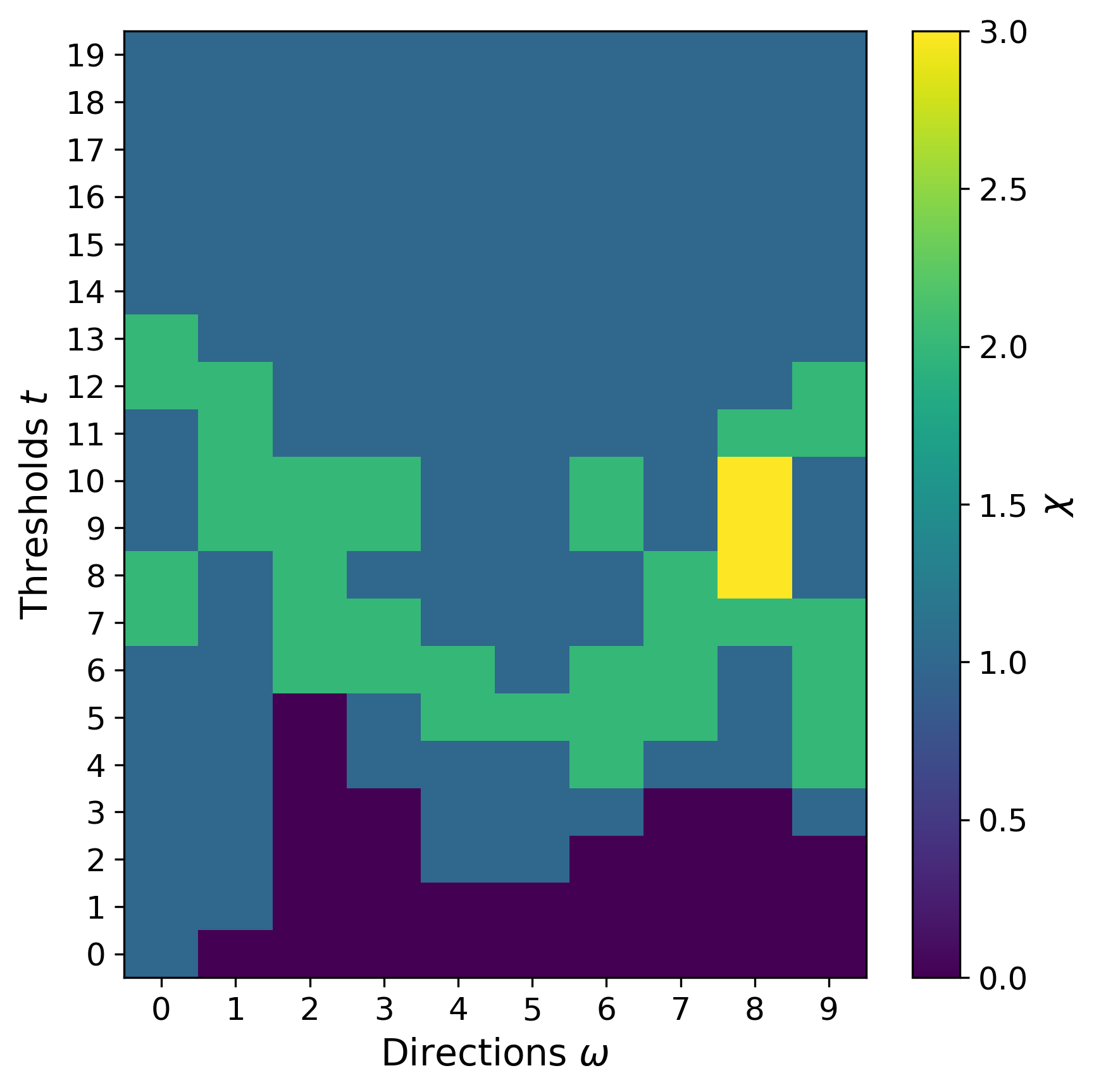}
    \caption{Euler Characteristic Transform}
    \label{fig:ect}
\end{subfigure}
\caption{\textbf{Extracting topological shape information from molecular graphs}. (a): Exemplary 2D graph representation of the acetic acid molecule, derived from its SMILES string \texttt{CC(O)=O}. The graph contains 8 vertices (2 carbon, 2 oxygen, and 4 hydrogen atoms) and 7 edges, resulting in an Euler characteristic of 1 ($\chi = 1$). (b): Filtration values of the nodes along the x-axis direction. (c) Filtration sequence along the x-axis direction. (d): ECC computed for the x-axis direction. In the figure, the x-axis represents the filtration thresholds values, and the y-axis shows the corresponding Euler characteristic values. (e): ECT example, built by stacking the ECC along 10 random directions from the molecule graph of (a). Columns corresponds to directions, rows to thresholds (set to 20), and color intensity encodes the Euler characteristic value.}
\label{fig:ectfrommolecule}
\end{figure}

In practice, molecules are often represented in formats that facilitate both human readability and algorithmic processing. One of the most widely used textual encodings is the SMILES (Simplified Molecular Input Line Entry System) notation, which encodes a molecule as a string describing the atoms and their connectivity through a series of characters and symbols. For example, the SMILES string \texttt{CC(O)=O} represents acetic acid. This format is compact, easily parsed, and widely supported in cheminformatics toolkits. However, SMILES do not directly convey geometric or spatial information, and small changes in the string can correspond to large structural differences. From molecular structures (e.g. SMILES, although alternative encodings exist), it is common to derive graph-based representations \cite{graphneuralnetworksbook}, where atoms are modeled as nodes and covalent bonds as edges, possibly enriched with additional features such as atomic types, bond orders, or aromaticity indicators. These molecular graphs serve as the foundation for numerous machine learning models, facilitating the use of graph-based algorithms and neural networks \cite{deng2022artificial}.
They also provide the basis for computing molecular representations such as fingerprints, descriptors, and our ECT-based representation. An example of the molecular graph of acetic acid is shown in Fig.  \ref{fig:moleculegraph}.

Traditionally, computational models for molecular machine learning tasks have relied on handcrafted molecular representations \cite{wigh2022review}. Among these are \emph{molecular descriptors}, which are numerical features derived from molecules, such as atom counts, topological indices, or electronic properties, and \emph{molecular fingerprints}, which represent the presence of specific substructures or chemical motifs (e.g., AVALON or MACCS) as binary or count vectors. Both types of features are computed directly from the molecular graph structure. These representations have been successfully applied in a variety of tasks such as virtual screening, quantitative structure–activity relationship (QSAR) modeling, molecule classification, and property prediction. However, they often suffer from a lack of expressiveness and poor generalization to out-of-distribution chemical spaces, especially in the presence of subtle variations in molecular geometry \cite{baptista2022evaluating,david2020molecular,xiong2019pushing}.

In recent years \emph{Graph Neural Networks} (GNNs) \cite{graphneuralnetworksbook}, have become a dominant paradigm in molecular property prediction, extending neural networks to graph-based data. The core mechanism of most GNNs is message passing, which iteratively propagates information across the graph elements. While GNNs can capture relational and structural information more flexibly than fixed descriptors, they often lack explicit access to multiscale shape information. Moreover, they may struggle to distinguish between molecules that are topologically or geometrically distinct but share similar local connectivity \cite{jiang2021could}.
Methods from \emph{Topological Data Analysis} (TDA) \cite{edelsbrunner2022computational} provide a complementary perspective. Instead of relying on raw atomic positions, handcrafted atomic features or purely local graph structures, TDA extracts global shape features from data, capturing relevant geometric information at multiple scales \cite{lum2013extracting}, making it thus particularly suitable for applications in the life sciences~\cite{Waibel22a}. A particularly relevant tool within this framework is the Euler Characteristic Transform (ECT) \cite{munch2025invitation,Rieck24a,Roell24a,Turner14a}, a method that combines ideas from algebraic topology with geometric data analysis. The ECT operates by \emph{filtering} a shape, such as the molecular graph, along multiple directions in the feature space, and computing the \emph{Euler characteristic} \cite{leinster2008euler}, denoted as $\chi$. This geometrical-topological quantity  encodes information about the the number of connected components, holes, and voids at each step of the filtration. Moreover, it is an \emph{invariant}, i.e., it will  remain unchanged under any smooth transformation applied to a shape. Put briefly, the Euler characteristic can be seen as a summary statistic of the shape of a graph or simplicial complex. To fully understand this tool and how molecules are represented as graphs, we next provide a more detailed introduction about graphs and simplicial complexes, serving as the initial points for computing the Euler characteristic and the ECT.

\paragraph{Graphs.} Graphs enable the modeling of real-world systems by focusing 
on \emph{dyadic relationships} between elements. Formally, a \emph{graph} $G=\left(V,E\right)$ consists of a finite set of vertices $V = \{v_1, v_2, \dots, v_n\}$, which represent entities (e.g. atoms in a molecule), and a set of edges $E \subseteq \{\{u, v\} \mid u, v \in V \text{ and } u \neq v\}$, which represent pairwise relationships (e.g., chemical bonds). In cheminformatics, graphs are the natural way to represent molecules, nodes corresponds to atoms, typically encoded as a feature vector
, and edges to chemical bonds. However, to extract richer geometric and topological information, we can generalize graphs into structures called \emph{simplicial complexes}. The graph of the acetid acid molecule shown in Fig. \ref{fig:moleculegraph} (visualized in $2D$) contains 8 vertices, which corresponds to 2 carbon atoms, 2 oxygen atoms, and 4 hydrogen atoms and 7 edges, with $\{H,O\}$ being an example of an edge between two vertices~(representing a chemical bond).

\paragraph{Simplicial complexes.} An (abstract) simplicial complex $K$ is a data structure for representing topological spaces, which generalizes a graph by permitting more than mere dyadic relations. It is defined as a family of sets~(\emph{simplices}) that is closed under taking subsets, meaning that if a set (like a triangle) is part of the complex, then so are all its faces (edges and vertices). More formally, a simplicial complex is obtained by a nested family of simplices, which are the elementary building blocks, for example: a $0$-simplex can be thought of as a point (vertex), a $1$-simplex as an edge, a $2$-simplex as a filled triangle, and a $3$-simplex as a filled tetrahedron. Each $k$-simplex has $k+1$ faces obtained by removing one of the vertices. For example, the graph of Fig. \ref{fig:moleculegraph} can also be seen as a $1$-dimensional simplicial complex with 8 $0$-simplices and 7 $1$-simplices.
The Euler characteristic is a key topological invariant of a simplicial complex~$K$, being defined as:
\begin{equation}
    \chi(K) = \sum^n_{k=0} (-1)^k |K^{(k)}|,
\end{equation}
where $|K^{(k)}|$ denotes the number~(cardinality)  of $k$-simplices in the simplicial complex~$K$.
Hence, the Euler characteristic
is an alternating sum of the number of simplices (elements) in each dimension. In the case of~(molecular) graphs, which only consist of $0$-simplices (vertices) and $1$-simplices (edges), this is reduced to:
\begin{equation}
    \chi = |V| - |E|.
\end{equation}
For example, the Euler characteristic of the acetic acid graph presented in Fig. \ref{fig:moleculegraph} is 1 ($8$ vertices, $7$ edges).

Taken on its own, the Euler characteristic lacks sufficient complexity to fully describe a shape, but if we think of computing it at different \emph{scales} (thresholds), considering we have a dynamic object, which grows in the number of their components (vertex, edges, etc.) across time, we may observe significant changes on it.
This leads to the concept of the Euler Characteristic Curve (ECC), which tracks how the topological complexity of a shape evolves over different scales. 
To formalize this dynamic view and to compute the Euler characteristic at different scales, we use the notion of a filtration, leading to a \emph{filtered simplicial complex}. More formally, a \emph{filtered simplicial complex} is a collection of subcomplexes $\{K(t) \mid t \in \mathbf{R}\}$ of a simplicial complex $K$ such that $K(t) \subseteq K(s)$ for $t < s$ and there exists $t_{\max} \in \mathbf{R}$ such that $K(t_{\max}) = K$. The \emph{filtration time} (or filtration value) of a simplex $\sigma \in K$ is the smallest $t$ such that $\sigma \in K(t)$.
To illustrate the concept of a filtration, consider the graph representation of the acetic acid molecule shown in Fig.~\ref{fig:filtrationvalues}. Each node (atom) is assigned a scalar value, which in this case corresponds to its projection onto a fixed direction, which corresponds to the x-axis direction for that example. These scalar values determine the filtration times of the nodes: a node enters the filtration at the time equal to its value. Formally, this means we construct a sequence of subgraphs (subcomplexes), $\{K(t) \mid t \in \mathbf{R}\}$ such that each $K(t)$ contains all nodes and edges whose filtration time is less than or equal to $t$. Since an edge can only appear after both its incident nodes have appeared, the filtration is nested, we have $K(t) \subseteq K(s)$ for $t < s$. An illustrative example is shown in Fig.~\ref{fig:filtrationsequence}, where the molecular graph of Fig.~\ref{fig:moleculegraph} evolves as the filtration parameter increases along the x-axis direction. At each step, new atoms (nodes) and bonds (edges) are incorporated according to their associated filtration values, as shown in Fig.~\ref{fig:filtrationvalues}, progressively revealing the full molecular topology.

\paragraph{Euler Characteristic Transforms.} While the ECC already provides a summary of how the topological complexity of a shape evolves with respect to a \emph{single} filtration parameter, it is often insufficient to fully characterize a high-dimensional or intricate structure like a molecule. This is where the \emph{Euler Characteristic Transform}~(ECT) comes into play, making it possible to characterize shapes based on multiple filtrations, parameterized using a \emph{direction vector}.
Specifically, each direction provides a different view of the data, capturing how the topological features of the structure unfold when the complex is filtered using a different criterion. To better understand this, imagine a 2D or 3D object~(like a molecule embedded in space). If we project the molecule along a certain direction, for example along the $x$-axis, we can define a filtration by sweeping a hyperplane orthogonally to that axis and including simplices as their associated values fall below a certain threshold. This process gives us an ECC in the $x$-direction. Now, if we repeat this process in another direction, for example along the $y$-axis, we will generally obtain a different ECC, since the structure of the molecule may appear differently from that angle. The key idea is that \emph{each} direction  typically provides complementary topological information.
Thus, the ECT is formally defined as the collection of Euler characteristic curves obtained by filtering a shape along a family of directions, which for molecule graphs are sampled  randomly in the high-dimensional node feature space.
For example, for a given molecule and its graph, the ECT produces a collection of curves (ECCs), each corresponding to a different direction in the node feature space. By aggregating all these curves into a single descriptor, the ECT captures how the topology of the molecule evolves as a function of spatial thresholds, providing a rich and compact descriptor of molecular shape. Refer to Fig.~\ref{fig:eccs} for the ECC of the molecule graph in Fig.~\ref{fig:moleculegraph} along the $x$-axis direction, and to Fig.~\ref{fig:ect} for an ECT example of this graph, built by stacking multiple ECCs, visualized as a matrix, where each column corresponds to a direction, each row to a threshold level in the filtration process, and the color intensity encodes the Euler characteristic value. Note that the numbers shown on the $x$-axis and $y$-axis of these figures do not represent the actual threshold and directions values used during filtration, but rather the index of the threshold and direction step.
The resulting feature set, given by the ECT, encodes both multiscale and directional information, offering a level of expressiveness that can be used in machine learning models as input features for both classification and regression tasks, and that can complement traditional descriptors and graph-based methods. Hence, the ECT has been successfully applied for various machine-learning tasks~\cite{ECTapplicationThreshold,Roell24a}.While the Euler characteristic itself has previously been used in molecular dynamics \cite{smith2023topological,laky2024fast}, to our knowledge, our work is the \emph{first} to explore the use of the ECT in the context of molecular machine learning.


\section{Materials and Methods}\label{sec:materials}

Having described a novel and multiscale topology-based approach to represent molecules, we conduct an extensive suite of experiments to explore and assess its effectiveness against established methods such as molecular fingerprints, descriptors, and graph neural networks.

\subsection{Datasets}

We concentrate on a number of datasets from different biochemical domains. 
Specifically, we use a collection of nine protein-ligand binding affinity datasets, all focused on predicting the inhibition constant $K_i$, a key measure of how strongly a molecule binds to a given protein target. These datasets are sourced from BindingDB\footnote[2]{\url{https://www.bindingdb.org/}} and cover a variety of biologically relevant proteins. The dataset associated with the alpha1A adrenergic receptor (ADRA1A) contains 1959 molecules; the one for the arachidonate 5-lipoxygenase-activating protein (ALOX5AP) includes 1636 molecules; and the dataset for the ataxia telangiectasia and Rad3-related protein (ATR) consists of 1411 molecules. For human dipeptidyl peptidase-4 (DPP4), the dataset comprises 3345 molecules, while the Janus kinase proteins JAK1 and JAK2 include 3058 and 1449 molecules, respectively. The kappa-type opioid receptor (KOR) dataset contains 1685 molecules, and the two datasets corresponding to subtypes of muscarinic receptors (MUSC1 and MUSC2) consist of 2129 and 1109 molecules, respectively.
Note that these molecule counts reflect the outcome of a preprocessing step in which only molecules with \emph{unique} SMILES representations were retained, and any entries with missing data were excluded. Although each dataset targets a different protein, they all share a unified prediction goal, namely the estimation of  $K_i$, reflecting molecular binding strength.
In addition, it has to be considered that the experiments have been carried out both for all datasets independently and for all datasets together. The combined dataset thus comprises a total of 16558 unique molecules.

\subsection{Methods}
We distinguish our description of the methods based on the type of molecular representation they use,
since this directly influences the type of machine learning methods we can apply.
Table~\ref{representations} lists the different representations, as well as their category and dimensionality~(i.e., the dimension of input features). Note that for graphs the concept of dimensionality as a feature vector is meaningless. In addition,  note that there is an existing descriptor called \textit{TOPO}\cite{mauri2020alvadesc}, which is based in topological information but not in a multiscale sense.

\begin{table}[tbp]
\centering
\caption{List of molecular representations, including their category and dimensionality, which refers to the corresponding feature vector dimensionality. They are grouped by the representation category and within that grouping sorted alphabetically.}\label{representations}
\renewcommand{\arraystretch}{0.2}
\begin{tabular}{{@{}llr@{}}}
\toprule
\textbf{Representation name} & \textbf{Representation Category} & \textbf{Dimension}\\
\midrule
Molecule graph & Graph & - \\
\midrule
AVALON & Fingerprint &  1024 \\
CATS2D & Fingerprint  &  189 \\
ECFP4 & Fingerprint &   1024\\
EState & Fingerprint &  79 \\
KR & Fingerprint &  4860 \\
MACCS & Fingerprint &  166 \\
MAP4 & Fingerprint &  1024 \\
Pharm2D & Fingerprint &  1024 \\
PubChem & Fingerprint &  881 \\
RDKit & Fingerprint &  1024 \\
\midrule
2DAP & Descriptor &  1596 \\
ConstIdx & Descriptor &  50 \\
FGCount & Descriptor&  153 \\
MolProp & Descriptor &  14 \\
RingDesc & Descriptor &  35 \\
TOPO &  Descriptor &  74 \\
WalkPath & Descriptor &  46 \\
\midrule
ECT (ours) & Multiscale Topological & 2528\\
ECT + Fingerprint (ours) & Multiscale Topological & 3552\\
\bottomrule
\end{tabular}
\end{table}

In total, we have 5 different categories of representations, the molecular graph, fingerprints, descriptors, the ECT, and the ECT + fingerprint. The latter two denote our two novel ECT-based approaches. We denote by ``ECT + fingerprint'' a representation consisting of the ECT combined with the AVALON molecular fingerprints~(via concatenation). This has been selected because the AVALON fingerprint is one the most commonly used fingerprints and generally performs well. In future work, we aim to study whether there are specific combinations of representations that can perform even better; please refer to our discussion in (Sec.~\ref{sec:conclusion}). 

When using the molecular graph representation, we use graph neural networks as the underlying machine learning models, since they are specialized for this type of data input. Concretely, we used two standard GNN models, a graph attention network (GAT) and a graph convolutional network (GCN), as well as an specialized graph neural network for molecular learning, named ``AttentiveFP''~\cite{xiong2019pushing}. For all other representations, we use an XGBoost~\cite{10.1145/2939672.2939785} model,
which has consistently shown strong performance in variety of tasks. 

\subsection{Experimental Setup}

Molecule graphs are extracted from the SMILES strings provided in the datasets using the \texttt{
PyTorch Geometric\footnote{\url{https://pytorch-geometric.readthedocs.io/}}} Python package, which encoded nodes (atoms) as a 9-dimensional handcrafted feature vector. We compute the ECT of the molecular graphs over this multi-dimensional feature space. Specifically, we sample random directions in the feature space to obtain the geometric and topological structure of the molecule data. Based on a prior sensitivity analysis, we fixed the number of directions and filtration thresholds to $158$ and $16$, respectively.  This analysis revealed that the number of directions has a substantially \emph{greater} impact on predictive performance than the number of filtration thresholds. For instance, increasing the number of directions from 20 to 30 led to noticeable improvements in model accuracy. However, performance gains plateaued at higher values, with little to no improvement observed between 180 and 200 directions. In contrast, varying the number of filtration thresholds had minimal effect on performance across a wide range of values. Therefore, we selected $158$ directions to ensure sufficient expressiveness while limiting the number of thresholds to $16$ to reduce computational cost without sacrificing predictive accuracy. The resulting ECT-based feature vector has a dimensionality of $2528$, as summarized in Table~\ref{representations}. All ECTs have been computed using the \texttt{DECT} Python package~\cite{Roell24a}.  

To fairly evaluate the predictive performance of both traditional vector-based and ECT-based methods, we designed two experimental setups: one using XGBoost for vector representations (fingerprints, descriptors, and ECT), and another using graph neural networks (GNNs) operating directly on the molecular graphs. For vector-based representations, we employed the XGBoost regressor \cite{10.1145/2939672.2939785}, trained with $1000$ estimators, a learning rate of $0.01$, and a maximum tree depth of $5$. For the graph-based models, we explored three architectures: Graph Convolutional Networks (GCN), Graph Attention Networks (GAT), and the ``AttentiveFP'' model. The GCN was configured with two convolutional layers and $64$ hidden channels. The GAT model used eight attention heads, each with eight hidden units. Both GCN and GAT architectures followed the design principles of Platonov et al.~\cite{platonov2023critical}, including a two-layer perceptron after each neighborhood aggregation step, skip connections, and layer normalization.  The ``AttentiveFP'' model was configured with $64$ hidden units, a single output channel, four message-passing layers, and two attention-based update steps. Dropout was set to $0.2$. All GNN models were trained for $100$ epochs using the ADAM optimizer with a learning rate of $10^{-2.5}$ and a weight decay of $10^{-5}$. A $10$-fold cross-validation strategy was applied for all the methods with shuffling and a fixed random seed to ensure reproducibility. Model evaluation was conducted using two metrics: root mean squared error (RMSE), and coefficient of determination (R2).
Note that this has been done separately for each dataset as well as for the combined dataset.

\section{Results}\label{sec:results}

Table~\ref{results} provides an overview of all results for the different representations for each dataset, as well as for all datasets combined into a single one. 
Overall, we observe that already the ECT on its own provides meaningful and effective features for molecular machine learning, outperforming all alternative methods on ADRA1A. In \emph{combination} with the AVALON molecular fingerprint, we observe generally improved baseline predictive performance in 5/9 datasets, as well as on the combined dataset. Thus, the ECT performs \emph{consistently well}, showing its potential for creating generalizable features.

\begin{figure}[h]
    \centering
    \includegraphics[width=\textwidth,height=0.6\textwidth]{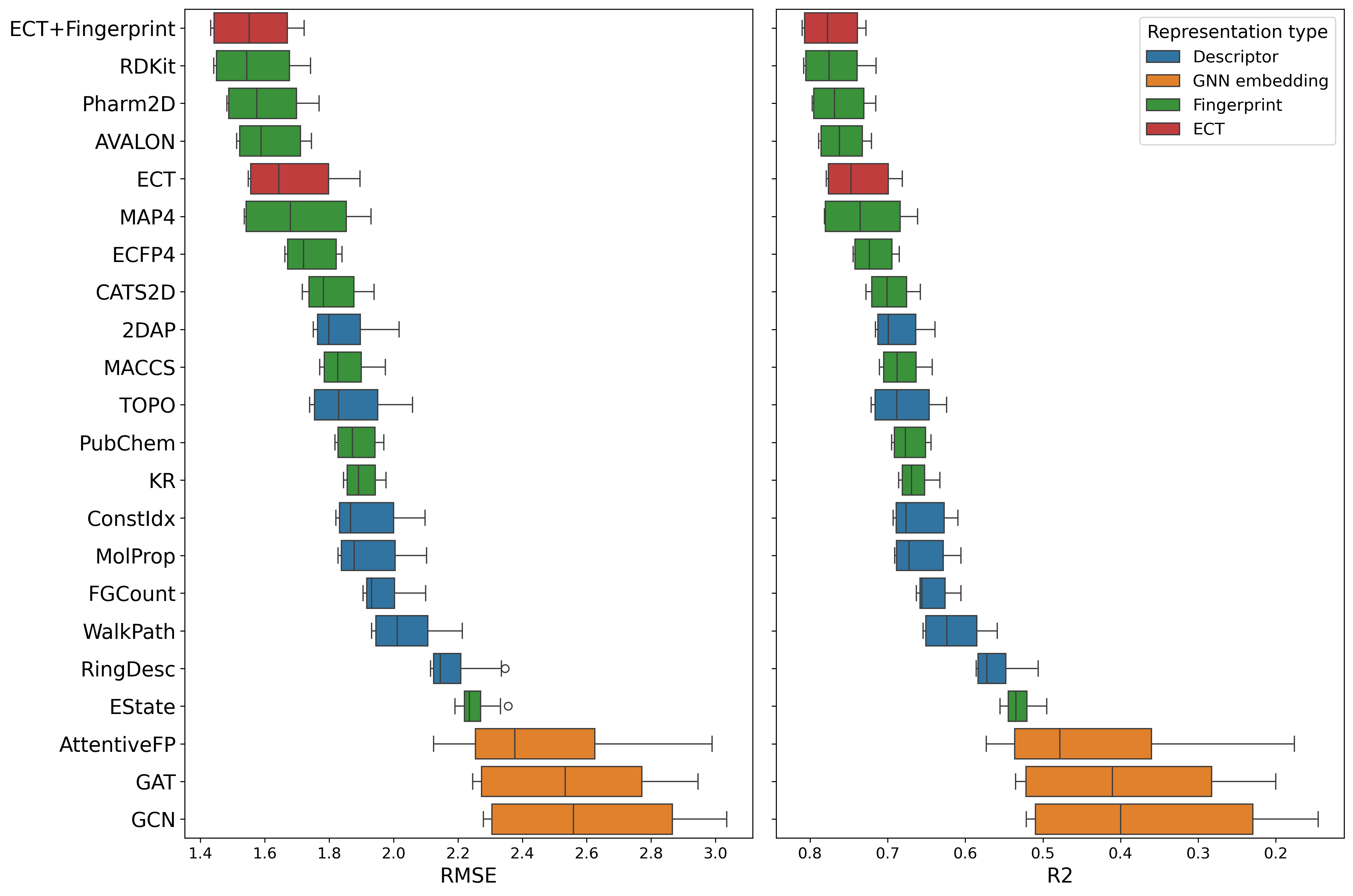}
    \caption{Boxplots showing the distribution of test errors for each representation method across all datasets. From left to right, the metrics displayed are RMSE and R2. Representations are ranked by their mean values: For RMSE,  \emph{lower} values indicate \emph{better} performance, while for R2, \emph{higher} values (closer to 1) indicate \emph{better} performance.}
    \label{fig:ComparisonBoxplotAllDatasetsTogether}
\end{figure}

Fig.~\ref{fig:ComparisonBoxplotAllDatasetsTogether} ranks the representations by their mean error for two different metrics (RMSE and R2). Performance is visualized using boxplots to show the error distribution for each representation method according the cross validation results. The representation combining our ECT-based representation and the AVALON fingerprint consistently achieves better values for RMSE and R2 metrics.
The low variance between cross-validation results is particularly noteworthy, showing the robustness of the approach.

\begin{table}[tbp] 
\centering
\caption{Results (RMSE, reported as mean ± standard deviation across 10-fold cross-validation) for the collection of molecular datasets. Rows corresponds to different representation methods grouped by category (GNNs, fingerprints, descriptors, and ECT-based). Columns correspond to datasets.}
\label{results}
\resizebox{\textwidth}{!}{
\renewcommand{\arraystretch}{1.3}

\sisetup{
  separate-uncertainty=true,
  table-format=1.2(2),
  detect-weight=true,
  detect-all = true,
}

\newrobustcmd\B{\DeclareFontSeriesDefault[rm]{bf}{b}\bfseries} 

\begin{tabular}{@{}l*{10}{S[table-format=1.2(2)]}@{}}
\toprule
\diagbox{Method}{Dataset} & {ADRA1A} & {ALOX5P} & {ATR} & {DPP4} & {JAK1} & {JAK2} & {KOR} & {MUSC1} & {MUSC2} & {Combined} \\
\midrule
AttentiveFP & 2.01 \pm 0.05 & 2.01 \pm 0.29 & 1.42 \pm 0.12 & 0.90 \pm 0.06 & 1.10 \pm 0.12 & 1.60 \pm 0.20 & 2.17 \pm 0.19 & 2.19 \pm 0.10 & 2.03 \pm 0.18 & 2.43 \pm 0.27 \\
GAT & 2.65 \pm 0.17 & 2.45 \pm 0.17 & 1.81 \pm 0.09 & 1.18 \pm 0.05 & 1.99 \pm 0.15 & 2.67 \pm 0.33 & 2.56 \pm 0.17 & 2.84 \pm 0.17 & 2.60 \pm 0.16 & 2.80 \pm 0.07 \\
GCN & 2.75 \pm 0.15 & 2.51 \pm 0.19 & 1.96 \pm 0.16 & 1.26 \pm 0.08 & 2.76 \pm 0.17 & 2.95 \pm 0.28 & 2.73 \pm 0.13 & 2.96 \pm 0.15 & 2.76 \pm 0.18 & 2.88 \pm 0.10 \\
\midrule
AVALON & 1.81 \pm 0.14 & 1.58 \pm 0.18 & 1.26 \pm 0.10 & 0.79 \pm 0.04 & 0.99 \pm 0.10 & \B 1.30 \pm 0.19 & 1.83 \pm 0.16 & 1.88 \pm 0.09 & 1.86 \pm 0.16 & 1.71 \pm 0.03 \\
CATS2D & 1.88 \pm 0.09 & 1.88 \pm 0.14 & 1.36 \pm 0.12 & 0.90 \pm 0.04 & 1.04 \pm 0.08 & 1.50 \pm 0.21 & 1.85 \pm 0.12 & 1.98 \pm 0.09 & 1.86 \pm 0.11 & 1.88 \pm 0.04 \\
ECFP4 & 1.81 \pm 0.11 & 1.71 \pm 0.17 & 1.30 \pm 0.09 & 0.84 \pm 0.04 & 1.01 \pm 0.12 & 1.31 \pm 0.18 & 1.90 \pm 0.14 & 1.98 \pm 0.06 & 1.83 \pm 0.13 & 1.81 \pm 0.03 \\
EState & 2.20 \pm 0.16 & 2.12 \pm 0.20 & 1.48 \pm 0.10 & 1.11 \pm 0.06 & 1.21 \pm 0.15 & 1.74 \pm 0.18 & 2.44 \pm 0.16 & 2.54 \pm 0.09 & 2.19 \pm 0.08 & 2.28 \pm 0.04 \\
KR & 1.82 \pm 0.14 & 1.78 \pm 0.20 & 1.36 \pm 0.09 & 0.88 \pm 0.04 & 1.03 \pm 0.12 & 1.42 \pm 0.19 & 1.91 \pm 0.18 & 2.00 \pm 0.07 & 1.85 \pm 0.11 & 1.95 \pm 0.02 \\
MACCS & 1.88 \pm 0.15 & 1.81 \pm 0.20 & 1.39 \pm 0.11 & 0.90 \pm 0.04 & 1.06 \pm 0.12 & 1.52 \pm 0.19 & 1.99 \pm 0.15 & 2.07 \pm 0.09 & 2.08 \pm 0.13 & 1.92 \pm 0.04 \\
MAP4 & 1.88 \pm 0.15 & 1.68 \pm 0.16 & 1.33 \pm 0.10 & 0.86 \pm 0.05 & 1.03 \pm 0.10 & 1.44 \pm 0.22 & 2.01 \pm 0.17 & 2.07 \pm 0.09 & 2.00 \pm 0.17 & 1.86 \pm 0.04 \\
Pharm2D & 1.80 \pm 0.11 & 1.68 \pm 0.21 & 1.35 \pm 0.10 & 0.79 \pm 0.05 & \B 0.98 \pm 0.12 & 1.47 \pm 0.20 & \B 1.81 \pm 0.11 & 1.89 \pm 0.12 & 1.79 \pm 0.13 & 1.71 \pm 0.03 \\
PubChem & 1.87 \pm 0.13 & 1.66 \pm 0.16 & 1.32 \pm 0.12 & 0.90 \pm 0.04 & 1.01 \pm 0.10 & 1.32 \pm 0.22 & 2.01 \pm 0.15 & 2.06 \pm 0.11 & 1.90 \pm 0.09 & 1.94 \pm 0.02 \\
RDKit & 1.83 \pm 0.13 & 1.56 \pm 0.13 & \B 1.26 \pm 0.10 & 0.78 \pm 0.05 & 0.99 \pm 0.11 & 1.36 \pm 0.15 & 1.85 \pm 0.13 & 1.89 \pm 0.08 & 1.91 \pm 0.13 & 1.69 \pm 0.03 \\
\midrule
2DAP & 1.84 \pm 0.14 & 1.66 \pm 0.13 & 1.33 \pm 0.11 & 0.90 \pm 0.05 & 1.02 \pm 0.10 & 1.45 \pm 0.23 & 1.91 \pm 0.14 & 1.98 \pm 0.10 & 1.88 \pm 0.08 & 1.91 \pm 0.05 \\
ConstIdx & 1.99 \pm 0.17 & 1.92 \pm 0.14 & 1.38 \pm 0.10 & 1.00 \pm 0.06 & 1.12 \pm 0.11 & 1.58 \pm 0.22 & 2.12 \pm 0.10 & 2.20 \pm 0.10 & 2.06 \pm 0.09 & 2.01 \pm 0.05 \\
FGCount & 1.93 \pm 0.16 & 1.71 \pm 0.16 & 1.36 \pm 0.10 & 0.91 \pm 0.06 & 1.01 \pm 0.08 & 1.40 \pm 0.17 & 1.98 \pm 0.16 & 2.09 \pm 0.07 & 1.87 \pm 0.15 & 2.01 \pm 0.06 \\
MolProp & 2.05 \pm 0.05 & 1.90 \pm 0.14 & 1.43 \pm 0.09 & 1.06 \pm 0.07 & 1.17 \pm 0.11 & 1.65 \pm 0.22 & 2.22 \pm 0.16 & 2.20 \pm 0.08 & 2.06 \pm 0.13 & 2.01 \pm 0.06 \\
RingDesc & 2.12 \pm 0.12 & 2.15 \pm 0.17 & 1.62 \pm 0.11 & 1.18 \pm 0.08 & 1.22 \pm 0.14 & 1.66 \pm 0.21 & 2.28 \pm 0.21 & 2.46 \pm 0.10 & 2.32 \pm 0.13 & 2.24 \pm 0.07 \\
TOPO & 1.89 \pm 0.13 & 1.92 \pm 0.13 & 1.50 \pm 0.08 & 1.04 \pm 0.06 & 1.20 \pm 0.09 & 1.64 \pm 0.18 & 2.01 \pm 0.16 & 2.10 \pm 0.11 & 1.92 \pm 0.09 & 1.97 \pm 0.05 \\
WalkPath & 1.97 \pm 0.13 & 1.98 \pm 0.13 & 1.42 \pm 0.10 & 1.06 \pm 0.06 & 1.16 \pm 0.12 & 1.61 \pm 0.21 & 2.16 \pm 0.19 & 2.17 \pm 0.09 & 2.09 \pm 0.09 & 2.13 \pm 0.05 \\
\midrule
ECT (ours) & \B 1.78 \pm 0.12 & 1.77 \pm 0.16 & 1.35 \pm 0.10 & 0.83 \pm 0.04 & 1.03 \pm 0.13 & 1.45 \pm 0.18 & 1.87 \pm 0.14 & 1.89 \pm 0.08 & 1.81 \pm 0.13 & 1.79 \pm 0.05 \\
ECT+ FP (ours) & \B 1.74 \pm 0.12 & \B 1.52 \pm 0.17 & 1.29 \pm 0.09 & \B 0.78 \pm 0.04 & 0.99 \pm 0.11 & 1.38 \pm 0.20 & 1.82 \pm 0.12 & \B 1.82 \pm 0.08 & \B 1.77 \pm 0.13 & \B 1.68 \pm 0.02 \\
\bottomrule
\end{tabular}

}

\end{table}

An overall observation is that most complex models~(like GNNs) that \emph{directly} take the molecular graph as its input tend to yield the \emph{worst} computational performance. This is in marked contrast to the other molecular representations~(based on XGBoost), where a single iteration of cross-validation takes around 10 to 20 seconds on a standard computer. In comparison, for GNNs, a single iteration takes about 40 to 60 seconds for GCN and GAT, and up to 100 seconds for ``AttentiveFP,'' which nevertheless leads to significantly worse predictive performance. This observation suggests that the use of deep learning models is not inherently justified by their~(computational and predictive) performance in this context. Rather, it appears that the choice of molecular representation plays a much more critical role in determining model effectiveness, highlighting that more complex architectures do not necessarily lead to better results. This observation is in line with other studies and might be attributed to the limited availability of data in the biomedical domain. 

\section{Conclusion}\label{sec:conclusion}

In this paper, we propose and implement an effective new molecular representation method based on the Euler Characteristic Transform (ECT), which captures multiscale topological information relevant to molecular property prediction tasks, specifically for predicting $K_i$, the inhibition constant. Our results on different datasets show that the ECT provides meaningful and effective features for molecular machine learning, and that its \emph{combination} with other approaches, specifically with the AVALON fingerprint, leads to improved predictive performance. This multiscale topological approach, which can capture shape characteristics across different resolutions, encodes structural information for learning tasks, which were hitherto not being considered or fully exploited by existing methods, presenting a promising avenue for molecular machine learning research.

\paragraph{Future work.} An exploration of \emph{how} the number of directions and thresholds used in the computation of the ECT affects its representational power is needed. We believe that identifying optimal or data-adaptive strategies for selecting these parameters will lead to more expressive and discriminative topological signatures, thus potentially improving performance in molecular property prediction tasks. This question is not only relevant for cheminformatics and biomedical applications, but could also benefit a wide range of domains where data can be modeled geometrically and topologically. In addition, we plan to extend and test the proposed approach in other molecular machine learning tasks. In this work, we tested a single hybrid strategy (ECT $+$ AVALON fingerprint) and observed improved predictive accuracy. Future work could
benchmark multiple hybrid strategies to identify synergies between multiscale topological
and traditional representations. Moreover, we aim to extend the ECT to molecular graphs with 3D positional information to quantify the geometric shape of the molecular structure itself, instead of, or in combination with, general atomic features. Further efforts will also include systematic hyperparameter tuning and the evaluation of different dataset splitting strategies, which can substantially influence both predictive performance and generalizability. Finally, our approach could also be extended  to other variants of the ECT, such as the Weighted Euler Characteristic Transform (WECT) \cite{jiang2020weighted} or the Smooth Euler Characteristic Transform (SECT) \cite{crawford2016functional}.

\paragraph{Data and Code Availability.}
All data and code used to generate the results presented in this manuscript are publicly available in a GitHub repository\footnote{\url{https://github.com/victosdur/ECTforMoleculeLearningTask}}.

\paragraph{Acknowledgements.}
This work was partially supported by REXASI-PRO H-EU project, call HORIZON-CL4-2021-HUMAN-01-01, Grant agreement ID: 101070028, by the Research Budget (VII PPIT-US) of the University of Seville, and by the Swiss State Secretariat for Education, Research, and Innovation (SERI).

\paragraph{Disclosure of interest.}
The authors have no competing interests to declare that are relevant to the content of this article.

%
%
%
\bibliographystyle{plain}
\bibliography{biblio}

\end{document}